# Application of Machine Learning in Early Recommendation of Cardiac Resynchronization Therapy

Brendan E. Odigwe
Computer Science & Engineering
U. of South Carolina
Columbia, SC 29208, USA
bodigwe@email.sc.edu

Francis G. Spinale
School of Medicine
U. of South Carolina
Columbia, SC 29208, USA
cvctrc@uscmed.sc.edu

Homayoun Valafar
Computer Science & Engineering
U. of South Carolina
Columbia, SC 29208, USA
homayoun@cse.sc.edu

*Abstract*— Heart failure (HF) is a leading cause of morbidity, mortality, and health care costs. Prolonged conduction through the myocardium can occur with HF, and a device-driven approach, termed cardiac resynchronization therapy (CRT), can improve left ventricular (LV) myocardial conduction patterns. While a functional benefit of CRT has been demonstrated, a large proportion of HF patients (30-50%) receiving CRT do not show sufficient improvement. Moreover, identifying HF patients that would benefit from CRT prospectively remains a clinical challenge. Accordingly, strategies to effectively predict those HF patients that would derive a functional benefit from CRT holds great medical and socio-economic importance. Thus, we used machine learning methods of classifying HF patients, namely Cluster Analysis, Decision Trees, and Artificial neural networks, to develop predictive models of individual outcomes following CRT. Clinical, functional, and biomarker data were collected in HF patients before and following CRT. A prospective 6-month endpoint of a reduction in LV volume was defined as a CRT response. Using this approach (418 responders, 412 non-responders), each with 56 parameters, we could classify HF patients based on their response to CRT with more than 95% success.

We have demonstrated that using machine learning approaches can identify HF patients with a high probability of a positive CRT response (95% accuracy), and of equal importance, identify those HF patients that would not derive a functional benefit from CRT. Developing this approach into a clinical algorithm to assist in clinical decision-making regarding the use of CRT in HF patients would potentially improve outcomes and reduce health care costs.

## I. Introduction

In the era of value-based healthcare, digital innovation, and big data, clinical decision support systems have become vital for organizations seeking to improve care delivery. Clinical decision support (CDS) tools can analyze large volumes of data and suggest the next steps for treatment, flagging potential problems and enhancing care team efficiency [1]. CDS tools can be used in disease management[2] or to aid clinicians with the administration of interventions[3]. At the core of many of these improved CDS tools are techniques and procedures that usually involve artificial intelligence/Machine learning of some sort[1].

Artificial intelligence (AI) based predictive tools such as Artificial Neural Networks (ANNs) [4], [5] have been implemented in recommender systems to enhance shopping experience [6], assist with speech recognition, and natural language processing [7]. They have also been incorporated into a plethora of other domains. Extensive efforts have been made to create new mechanisms of predictive modeling (such as LSTMs, DNNs, and CNNs) [8], [9] as well as to improve already existing methods (such as decision trees). While ANNs have advanced substantially over the last few years, their impact and integration in medical diagnostics have been minimal. The lack of integration of machine learning techniques in Clinical Decision Systems and medical science is due to the challenging nature of analyzing medical data and the absence of well-annotated and relevant data appropriate for machine learning endeavors [10].

Despite these challenges, ANNs have been applied in the domain of healthcare in applications ranging from medical fraud detection [11] to the prediction of a patient's response to medical treatment [12] [13], thereby paving the way for extensive efforts towards personalized medicine. ANNs have been used in other diagnostic applications that include analysis of medical images [14], cancer detection [15]–[17], and ophthalmic disease [18], etc. More recently, ANNs have improved patient outcomes by optimizing personalized medical care [12]. In this regard, ANNs hold the potential to predict, and therefore, improve diagnosis and treatment strategies

Heart failure (HF) is a constellation of clinical signs and symptoms due to underlying defects in left ventricular (LV) function. While pharmacological therapy forms the mainstay for HF treatment, device-driven approaches such as Cardiac Resynchronization Therapy (CRT) have also become a standard of care.[19]–[21] The basis of CRT is to improve LV myocardial conduction patterns in HF patients. Clinical trials have shown that CRT can improve HF symptoms and functional outcomes in HF patients [19], [20], [22], [23]. However, these studies and retrospective reviews of CRT have identified that a large proportion of HF patients (30-50%) do not demonstrate an improvement in functional/clinical outcomes.[19], [22]–[24] Specifically, while HF patients may meet the criteria for CRT, such as reduced LV ejection fraction and prolonged myocardial conduction, a large portion of these patients derive no benefit. Therefore, developing strategies to identify HF patients with the most significant probability of responding to CRT and conversely identifying a subset of HF patients with a low likelihood of CRT benefit holds significance in improving HF prognosis, treatment, and medical resource utilization.

Since CRT has relative effective rates of 50-70%, then a robust group of CRT "responders" and "non-responders" exist [19], [20], [22]–[24]. As such, we hypothesize machine learning approaches would be amenable for the prediction of CRT response. For this study, we utilized an existing CRT database reported previously, [22], [25] in which prospective classification of CRT response was established; and demographic, LV function, and biomarker measurements were made before CRT placement and at six months after CRT. We pursued machine learning methods of classifying candidate patients into responders and non-responders to identify those HF patients with the greatest probability of functional improvement with CRT.

## II. BACKGROUND

In medical decision-making (classification, diagnosing, etc.), there are many situations where decisions must be made effectively and reliably. Conceptual simple decision-making models with the possibility of automatic learning are the most appropriate for performing such tasks [26].

### A. Previous and Related Work

Classification is a task that requires the use of machine learning algorithms that learn how to assign a class label to examples from the problem domain [27]. Binary classification problems [28] aim to observe one of two categories by analyzing a series of attributes. An example is a medical diagnosis for a single medical condition (e.g., disease vs. no disease) based on a battery of tests. Numerous approaches to classification have been developed [29][30]. Still, the choice of technique to apply usually depends on the domain/task to which it is applied and the nature of the data. Classification algorithms have seen increased application in the medical domain in diagnosis[31], prognosis[32] as well as treatment [12].

While several large clinical trials have demonstrated that CRT can be beneficial to a cohort of HF patients, few have examined predictive algorithms. One of the more cited examples is the Multicenter Automatic Defibrillator Implantation Trial with Cardiac Resynchronization Therapy (MADIT-CRT- [23]). This study developed a composite clinical scoring system and related this to CRT functional outcomes. While more of a generalized classification scheme and not patient-specific, this was the first approach to develop a relative risk profile for HF patients and relation to CRT response. The SmartDelay Determined AV Optimization: A Comparison to Other AV Delay Methods Used in CRT [22] was a large clinical trial that used a prespecified quantifiable outcome measure at six months - LV volume reduction and collected demographic, functional, and biomarker profiling data at baseline (pre-CRT placement). One outcome from the SMART-AV study was identifying potential biomarkers (using a peripheral blood sample) that could be integrated into a clinical algorithm, such as the MADIT-CRT score.[25] This past study highlighted the complexity of developing modeling algorithms using standard parametric statistical approaches to predict a CRT response.

While these past studies highlighted the need for prediction algorithms in HF patients being considered for CRT, this remains an unmet medical need. Improper selection of HF treatment impact both patient health outcome and economics of the disease. Studies focused on the economic implications of CRT report [36], [37] an estimated value of $53,589/quality-adjusted life-year (QALY) gained over the patient's lifetime for those who respond positively to CRT therapy. However, equally important are those HF patients who undergo device placement and the costs associated with this treatment that yield no benefit.[37] An initial study reported using machine learning in CRT but only used a small number of clinical variables.[38]

Accordingly, the goal of this study was to expand the scope using multiple variable domains (demographic, functional, biomarker) from the past SMART-AV study [25] using machine learning approaches; namely, cluster analysis, decision trees, and ANNs to classify patients based on their expected outcome to CRT. This will be the first study to encompass a large number of variables obtained in HF patients before CRT and using these approaches to predict a prespecified functional outcome at 6 months following the institution of CRT.

### B. Description of the Data

The dataset was obtained from the SMART-AV clinical trial, for which all methods, approaches, and manner in which measurements were made have been published[34]. The entire dataset contained 1045 patients with 3 observations (for each patient) recorded at months

0, 3, and 6. Each observation consisted of 80 attributes or parameters. The "response" parameter indicated the patient response to CRT at the end of the 6-month observational period. The investigations and results discussed in this paper have been performed using only data observations obtained in month 0 (at baseline).

Patients with more than 10 illegitimate parameter values (missing values) were eliminated from our investigation, resulting in 830 remaining patient records. 418 of these patients were classified as responders, while 412 were classified as non-responders at the end of the 6 months of CRT.

From the 80 parameters recorded for all patients, some were considered preliminarily unnecessary and were therefore removed based on the following criteria:
- Those considered to be commentary, e.g., Patient number, Patient Status, Group, etc. This accounted for 18 parameters.
- Those observed to be constant values across all patients regardless of response classification. This accounted for 3 parameters.
- Parameters (values) observed to have been repeated (with a correlation coefficient of 1) were eliminated. This accounted for 3 parameters.

This effectively reduced the parameters used for computational experimentation to 56 (55 input parameters and 1 output).

Missing or corrupted data points (NAN) were replaced with the mean across the population cohort to avoid skewing the results of the experiments. The mean is assured to be within the standard value range for all features and a reasonable estimate for the missing values.

Dataset values were normalized to accommodate substantially different unit ranges of various parameters. For instance, the patient's age is measured in days (range was in thousands), while other parameters, usually categorical, have single-digit values. This disparity in the range of data is automatically resolved by normalization of the dataset during the training/testing processes. The ground truth classification of the subjects in the dataset was assessed and established by clinicians before the experimentation described in this paper.

III. MATERIALS AND METHODS

We investigated the use of computational methods, particularly Cluster Analysis, Decision Trees, and Artificial Neural Networks, for classifying candidate patients as responders and non-responders to identify those HF patients with the highest probability of functional improvement with CRT. Our choice of the machine learning technique to utilize as predictive tools is in line with following the minimalist approach of employing the simplest model to satisfy the problem requirements and then extend efforts when more robust techniques are required.

A. *Cluster Analysis*

Cluster analysis is a class of techniques used to classify objects or cases into relative groups called clusters without any information from a supervisor. This class of ML is, therefore, one of the most prevalently used unsupervised learning techniques. Cluster analysis is a task of grouping a set of things so that objects in the same group are more similar to each other than to those in other groups based on some metric of similarity. In cluster analysis, there is no prior information about the group or cluster membership (therefore unsupervised learning) for any of the objects [40]. It is considered one of the more common unsupervised learning techniques.

Cluster analysis involves formulating a problem, selecting a distance measure, selecting a clustering procedure, deciding the number of clusters, interpreting the profile clusters, and finally, assessing the validity of clustering, all of which were performed during our investigations. We performed K-Means clustering and Hierarchical clustering on the patient data in our dataset.

K-Means stores "k" centroids that it uses to define clusters. A point is considered to be in a particular cluster if it is closer (based on some measure of distance) to that cluster's centroid than any other centroid. It finds the best centroids by alternating between assigning data points to clusters based on the current centroids and choosing centroids (points that are the center of a cluster) based on the current assignment of data points to clusters [41].

Hierarchical clustering is an algorithm that groups similar objects into a set of clusters, where each cluster is distinct from the other cluster, and the objects within each cluster are broadly similar to each other [42].

B. *Decision Trees*

Decision trees (DT) are a reliable and effective decision-making technique that provides high classification accuracy with a simple representation of gathered knowledge. They have been used in different areas of medical decision-making [43]. They learn from the provided data to produce an approximation with a set of if-then-else rules. The deeper the tree, the more complex the decision rules, and the fitter the model. DT breaks down a dataset into increasingly smaller subsets while, at the same time, an associated decision tree is incrementally developed[44]. Decision trees use different object attributes to classify different subsets of objects. Their great advantage is that they do not use a fixed number of predetermined features. In the decision tree approach, the members of a set of objects are classified as either positive or negative instances (in our case, patients that responded positively to CRT and patients that did not). Candidate attributes that may describe the concept are then outlined [45].

A decision tree construction tool uses outlined attributes to formulate the appropriate decision tree that identifies all positive instances of the underlying concept according to subjects with known classification. The first set of objects used for tree generation is usually called the training set. This initial decision tree becomes a basis for:
1. Forecasting whether a new subject (previously unseen) is a positive or negative instance of the concept being modeled.
2. Investigating the hierarchical representation of the most critical attributes. [46]

As the DT is incrementally developed and the dataset is broken down, class entropy is calculated at each decision to determine if maximal information has been gained from the decisions made on our subjects. The subjects, in this case, are the patients. Our primary interest is to predetermine patients' responses and to identify the contributing attributes supporting each decision. A restriction on the minimum number of patients at a decision node was implemented to prevent over-training. Parameters used for this investigation included patient attributes such as left ventricular end-systolic volume, ischemic cardiomyopathy, systolic and diastolic blood pressure, resting heart rate, weight, body mass index, age, etc.

*C. Artificial Neural Networks*

Our investigations utilized Artificial Neural Networks [5](ANNs) as another predictive machine learning approach. ANNs are brain-inspired systems, which are intended to replicate the way that humans learn. Neural networks consist of input and output layers and (in most cases) a hidden layer consisting of units that transform the input into complex abstracted information that the output layer can use. They are excellent tools for finding patterns that are far too complex or numerous for a human analyst to extract.

We utilized several ANN architectures with varying complexities, including the legacy shallow ANNs [47]. Our approach involved rapid development and experimentation libraries such as TensorFlow[48] and Keras [49].

*D. Evaluation Process*

We set out to classify CRT candidate patients into one of two classes: Responders (Positive responses) and Non-responders (Negative responses).

We repeated the experiments multiple times while exploring different partitioning of the training and testing population with no changes to the detection mechanism. These results were averaged to obtain a fair performance evaluation of the computational approach. For each experiment (except for cluster analysis), we separated the dataset in a 70-30 ratio, i.e., 70% for training and 30% for testing. This distribution amounted to about 581 patients for training and 249 patients for testing. The members of each set were randomly assigned each time; therefore, the class composition changed but still maintained relatively equal distribution due to stratification. For each experiment, we obtained accuracy values for each class separately, and then we used these metrics to evaluate the performance of the model as a whole. Some of these metrics involved keeping count of;

- **Sensitivity** – the rate at which Responders are accurately classified as such (shown in eq 1).
- **Specificity** – the rate at which Non-Responders are accurately classified as such (shown in eq 2).
- **Accuracy** – the total number of correct classifications over the entire data set (eq 3).

$$Se = \frac{TP}{TP+FN} \quad (1)$$

$$Sp = \frac{TN}{TN+FP} \quad (2)$$

$$Acc = \frac{TP+TN}{TP+FP+TN+FN} \quad (3)$$

IV. RESULTS AND DISCUSSION

*A. Cluster Analysis*

Our initial step in understanding the complexity of this task consisted of performing cluster analysis. This exercise aims to separate the patients based on the natural separation of the data in some hyperdimensional space and compare the similarity of patient clusters to that of the expert-based classification. The corroboration between the two approaches could be interpreted as a successful outcome. We performed categorical clustering and K-Means clustering of the patient data. Our utility of K-Means used Euclidian distance as the measure of similarity during neighbor identification and cluster assignments. Unlike K-Means, hierarchical clustering permits the use of different distance measures such as Euclidian [50], Manhattan [51], and Cosine [52]. We performed hierarchical clustering using all three distance measures, which produced accuracies of 64%, 36%, and 64%, respectively. Based on this outcome, we used Euclidian distance for all subsequent experiments. The comparative results between the Hierarchical Clustering and K-Means Clustering are shown in Table 1.

With this approach, we found that patients could not be trivially separated into response classes as investigations yielded maximum classification accuracy of 57% and 64% for K-Means and Hierarchical clustering, respectively, while employing Euclidian distance as the distance metric (Table 1). It is clear that the clustering algorithms used are inadequate for classifying the patients. Therefore, more robust classification techniques were explored to address the problem better. Figure 1 illustrates the most successful dendrogram that was obtained from Hierarchical Clustering using the Euclidian distance metric. Based on this information, K-Means and Hierarchical clustering techniques were concluded to be insufficient in accurately classifying the patients in this task.

*Table 1: Table of Cluster Analysis Performance*

| Method | Accuracy | Sensitivity | Specificity |
|---|---|---|---|
| Hierarchical Clustering | 64% | 75% | 52% |
| K-Means Clustering | 57% | 56% | 58% |

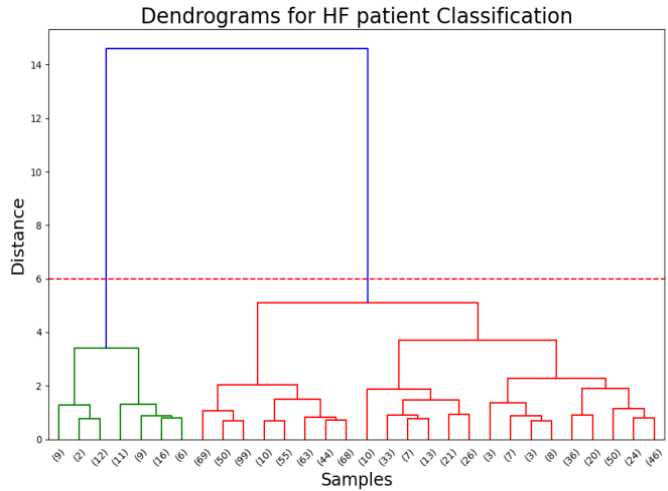

*Figure 1: Plot for Hierarchical Clustering of HF Patients. Red – Responders, Green – Non-Responders (dendrogram has been truncated before plotting with values on the "samples" axis detailing the number of samples already combined)*

B. *Decision Trees*

The HF data was used to develop Decision Trees with the desired performance and attributes. When unconstrained, decision trees can over-develop to achieve a 100% performance on the training set. To avoid overtraining, the resultant tree can be pruned by specifying the minimum number of samples a leaf-node was permitted to have, restricting the maximum size of a tree. Pruning the tree structure is also a way to prevent overfitting because the generated decisions for classification would not be overly personalized and therefore do not generalize to a broader population of patients. Our experiments started by allowing maximum tree growth (minimum samples per leaf-node of 1) and then gradually increasing the minimum samples per leaf until the performance was no longer acceptable. The acceptable tree performance was chosen to be an accuracy of no less than 85%. Figure 2 illustrates the performance of the decision trees as a function of the minimum population size of the leaves. Based on these results, a pruned tree with leaf-node population size of 20 was selected for further evaluation.

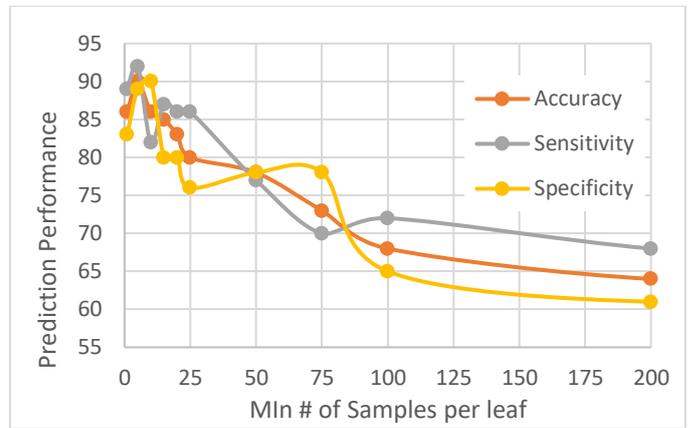

*Figure 2: Plot of Decision Tree performance with varying Minimum # of samples per node*

Using the chosen decision tree structure and restriction criteria, we evaluated the performance of the tree by randomly partitioning our data into 80% training patterns and 20% testing patterns. This process was repeated five times to obtain averaged fair values for performance. The decision tree classification experiment yielded results shown in Table 2, exhibiting an average performance of 87.2% in correctly predicting the patients' response to the CRT therapy. Average specificity (the percentage of responders who were classified as responders) was 87.4%, and average sensitivity (the rate of non-responders classified as non-responders) was 86.8%.

*Table 2: Decision Tree Classification %Performance*

| Exp # | Accuracy | Specificity | Sensitivity |
|---|---|---|---|
| 1 | 89 | 86 | 91 |
| 2 | 85.5 | 89 | 82 |
| 3 | 90 | 89 | 91 |
| 4 | 86 | 84 | 88 |
| 5 | 85.5 | 89 | 82 |
| Avg | 87.2 | 87.4 | 86.8 |

C. *Artificial Neural Networks*

Multiple neural network models were developed and incrementally modified to obtain a network architecture that produced the optimal classification accuracy. We began with a single-layer perceptron and gradually increased the number of neurons and hidden layers while observing the changes in classification accuracy. Previous research in deep learning [47] suggests that most problems can be solved with 1-2 hidden layers, in which the number of neurons in each hidden layer is likely between the number of neurons in the previous and subsequent layers. We performed experiments using a single hidden layer starting from 5 neurons and increasing the number of neurons in steps of 5 until the number of neurons was approximately equal to the number of input neurons. The performance of each single-hidden layer architecture is shown in Figure 3. With a performance limited to less than 76%, a second hidden layer was added, and experimentation continued.

While investigating an appropriate number of neurons for the second hidden layer, we intuitively chose 4 different values for the number of neurons in the first hidden layer (10, 20, 30, and 40). Our exploration of the size of the second hidden layer consisted of starting with two neurons and increasing in steps of 2 until no more improvement was observed in the performance. The impact of growing the second hidden layer on performance was observed to obtain the most efficient network architecture (i.e., the smallest network that produced acceptable results). Figures 4-7 illustrate the varying performance of a two hidden layer network as a function of network size.

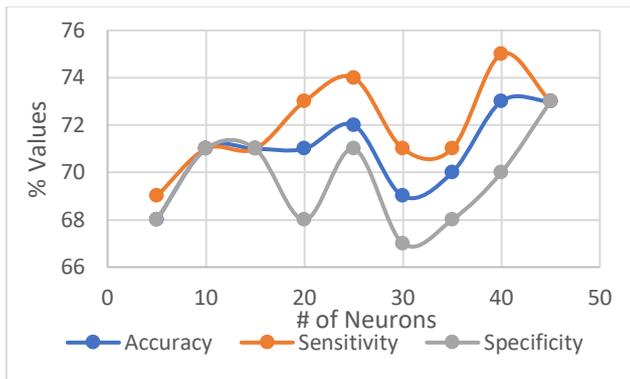

*Figure 3. The performance of a single-layer network as a function of varying hidden neurons.*

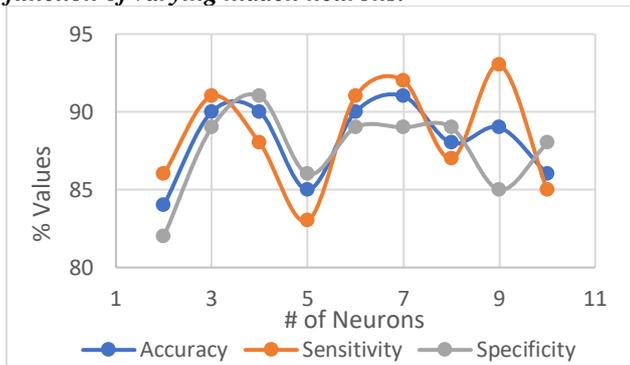

*Figure 4. The performance of a two hidden-layer network as a function of the second-layer size. The first layer is fixed at 10 hidden neurons.*

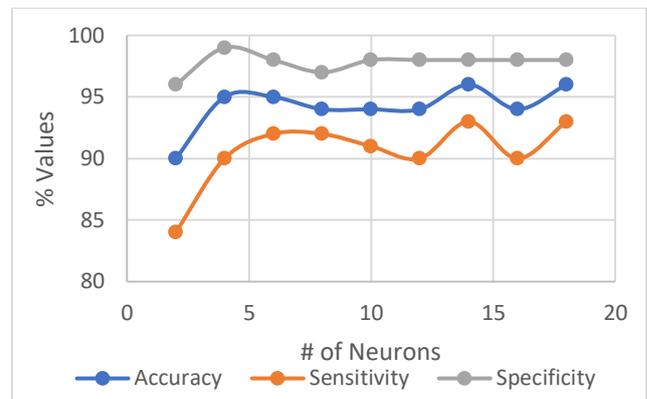

*Figure 5. The performance of a two hidden-layer network as a function of the second-layer size. The first layer is fixed at 20 hidden neurons.*

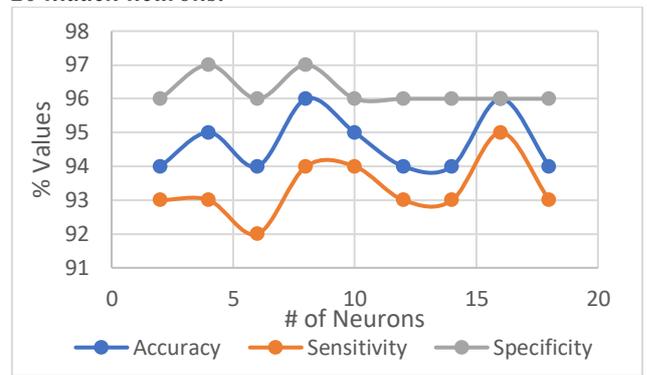

*Figure 6. The performance of a two hidden-layer network as a function of the second-layer size. The first layer is fixed at 30 hidden neurons.*

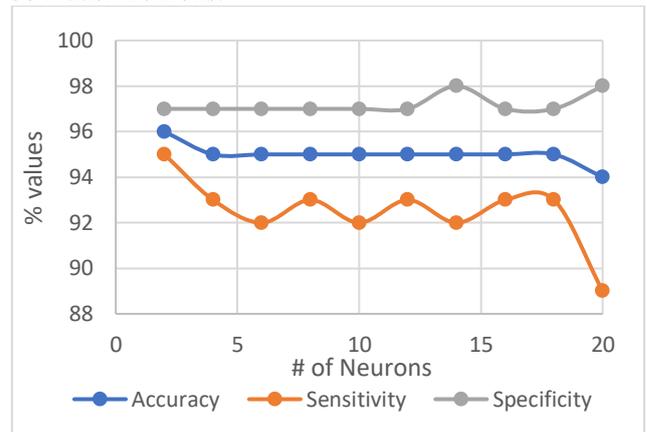

*Figure 7. The performance of a two hidden-layer network as a function of the second-layer size. The first layer is fixed at 40 hidden neurons*

Based on the results shown in these figures, the choice for the number of first layer neurons is 20, and from the results when experimenting with 20 neurons in the first layer, the highest performance was obtained with 14 neurons in the second hidden layer. Therefore, for the sake of our investigations, the optimal neural network architecture was; **53 -> 20 -> 14 -> 1** (53 input neurons, 20 neurons in the first hidden layer, 14 neurons in the second hidden layer, and one neuron in the output layer).

The ANN experiments were repeated five times using this network architecture, sharing the dataset into 70% for training and 30% for testing. Since this is a binary classification problem, we employed the sigmoid activation function in the output layer and linear activation functions for each preceding layer of the network.

The classification experiment yielded results showing that 95.2% of the participating patients were accurately predicted into their response classes. Specificity (the percentage of responders who were classified as responders) was 95.2%, and sensitivity (the rate of non-responders who were classified to be non-responders) was

95.2% (see Table 4). In general, the performance of this network indicated that there was little to no bias towards the prediction of responders or non-responders, further bolstering the integrity of our classification procedure.

*Table 3: HF Patient Classification Performance with ANNs*

| Exp# | % acc for 1's | % acc for 0's | Pred Acc |
|---|---|---|---|
| 1 | 94 | 93 | 94 |
| 2 | 93 | 96 | 94 |
| 3 | 98 | 94 | 96 |
| 4 | 96 | 96 | 96 |
| 5 | 95 | 97 | 96 |
| Avg | 95.2 | 95.2 | 95.2 |

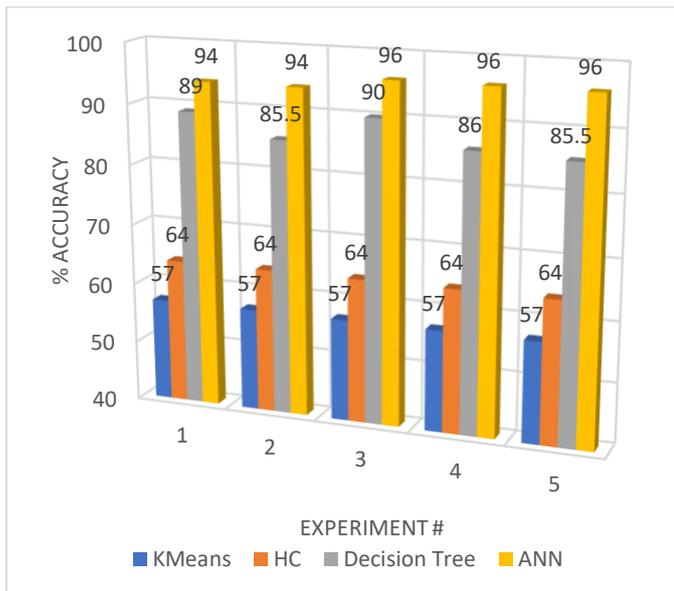

*Figure 8: Comparison of Classification Performance Between the Approaches Employed.*

## V. CONCLUSION AND FUTURE WORK

The advancements made in the field of machine learning make ML techniques suitable for use in the development of tools that aid in the medical decision-making process even though, for various justified reasons, their impact and integration in the field of medical diagnostics have been minimal [26], [31].

Machine learning models can recognize patterns and complex relationships between parameters that are not easily discernible by man. The parameter values used for this experiment were obtained from patients undergoing CRT.

We have demonstrated that we can (with 87% accuracy) identify HF patients likely to respond positively to CRT by following the sets of rules/questions produced by the decision tree. We have also demonstrated that a classification accuracy value of up to 95% is achievable with the employment of artificial neural networks.

Regardless of how the output classes are categorized, training ANNs to accurately distinguish between different classes of patients is valuable to the medical community. This experiment shows that ANNs are capable of exploiting relationships with and between medical data parameters. This approach and others like it will positively influence medical decision-making, administration of intervention procedures, and improves the practice of precision medicine.

The decision tree approach and the rules generated produce a means of prioritizing patient data parameters and presents us with the opportunity to extend our investigation efforts into dataset dimension reduction, where the highest priority parameters produced by the decision trees are used for neural network experimentation.

### ACKNOWLEDGMENT

This work was supported by the National Institute of Health grant HL095608 (FGS), a Merit Award from the Veterans' Affairs Health Administration (FGS), an unrestricted grant from Boston Scientific (FGS), and NIH P20… (HV).